%% file: main.tex
\documentclass[10pt,twocolumn,letterpaper]{article}

\usepackage[pagenumbers]{cvpr} 


\usepackage{microtype}
\usepackage[sort&compress,numbers]{natbib}
\usepackage{mathrsfs}
\usepackage{amsmath}
\usepackage{bm}
\usepackage{colortbl}
\usepackage{multirow}
\usepackage{adjustbox}
\usepackage{graphicx}
\usepackage{colortbl}
\usepackage{arydshln}
\usepackage{microtype} 
\usepackage{amsmath, amssymb, mathtools}
\usepackage{appendix}
\usepackage{multirow}
\usepackage{wrapfig}
\usepackage{multirow}
\definecolor{mygray}{gray}{0.6}
\definecolor{cvprblue}{rgb}{0.21,0.49,0.74}
\definecolor{mygray}{gray}{0.6}

\newcommand\nnfootnote[1]{%
  \begin{NoHyper}
  \renewcommand\thefootnote{}\footnote{#1}%
  \addtocounter{footnote}{-1}%
  \end{NoHyper}
}

\definecolor{cvprblue}{rgb}{0.21,0.49,0.74}
\usepackage[pagebackref,breaklinks,colorlinks,allcolors=cvprblue]{hyperref}

\def\paperID{10478} 
\def\confName{CVPR}
\def\confYear{2025}

\title{Treat Visual Tokens as Text? But Your MLLM Only Needs Fewer Efforts to See}

\author{~\textsuperscript{$\star$}Zeliang Zhang$^{1}$, ~\textsuperscript{$\star$}Phu Pham$^{2}$ , \textsuperscript{$\star$}Wentian Zhao$^{3}$\textsuperscript{$\dagger$}, \textsuperscript{$\star$}Kun Wan$^{3}$\textsuperscript{$\dagger$}, Yu-Jhe Li$^{3}$,  Jianing Zhou$^{4}$, \\
Daniel Miranda$^{3}$, {Ajinkya Kale}$^{3}$, {Chenliang Xu}$^{1}$ \\
  $^1$University of Rochester \quad \quad $^2$Purdue University \quad \quad $^3$Adobe Inc. \quad \quad $^4$ UIUC\\
  \texttt{ \{zeliang.zhang, chenliang.xu\}@rochester.edu}, 
  \texttt{phupham@purdue.edu} \\
  \texttt{\{wezhao, kuwan, jhel, miranda, akale\}@adobe.com}\\
  \texttt{zjn1746@illinois.edu}
}

\begin{document}
\maketitle


\begin{abstract}
\nnfootnote{$\star$ indicates the equal contribution with random order. } 
\nnfootnote{$\dagger$ indicates the project leader. }
By treating visual tokens from visual encoders as text tokens, Multimodal Large Language Models (MLLMs) have achieved remarkable progress across diverse visual understanding tasks, leveraging the robust architectures of Large Language Models (LLMs). However, as token counts grow, the quadratic scaling of computation in LLMs introduces a significant efficiency bottleneck, impeding further scalability. Although recent approaches have explored pruning visual tokens or employing lighter LLM architectures, the computational overhead from an increasing number of visual tokens remains a substantial challenge. \textcolor{black}{It also raises challenges to decide important visual tokens for different sample, which introduces another computation overhead.}  

In this study, we investigate the redundancy in visual computation at both the parameter and computational pattern levels within LLaVA, a representative MLLM, and introduce a suite of streamlined strategies to enhance efficiency. These include neighbor-aware visual token attention, pruning of inactive visual attention heads, and selective layer dropping for visual computations. By implementing these strategies in LLaVA, we achieve a reduction in computational demands of $88\%$ while maintaining model performance across key benchmarks. Additionally, we validate the existence of visual computational redundancy in other MLLMs, such as Qwen2-VL-7B and InternVL-2.0-4B/8B/26B. These results present a novel pathway for MLLMs to handle dense visual tokens with minimal computational costs. \textcolor{black}{Our pruning method only requires one time to discover the important visual-related computation and is sample-agnostic. This indicates that, \textit{You Only need Pune Once (YOPO)} to accelerate your MLLMs.} Code and model checkpoints are available at \url{https://github.com/ZhangAIPI/YOPO_MLLM_Pruning}.
\end{abstract}

\section{Introduction}

After the tremendous success of LLMs~\cite{achiam2023gpt,team2023gemini, touvron2023llama, jiang2023mistral}, researchers began exploring the introduction of additional modalities, such as images~\cite{alayrac2022flamingo, li2023blip, llava-1.0, internvl}, videos~\cite{video-llava}, and audio~\cite{audio-qwen}, into this architecture. Over the past two years, MLLMs, which incorporate more modalities, have been continuously improving performance. MLLMs based on the LLMs architecture have gained cross-modal understanding capabilities without significantly compromising the original abilities of LLMs. To further enhance the capabilities of MLLMs, various efforts are being made simultaneously, including architectural improvements~\cite{mckinzie2024mm1,wang2023cogvlm,tong2024cambrian,tong2024eyes}, data curation~\cite{changpinyo2021conceptual,sharma2018conceptual,chen2023sharegpt4v,chen2024allava}, cross-modal alignment~\cite{zhang2023llama-adapter,alayrac2022flamingo,wang2023cogvlm,mckinzie2024mm1}, and training recipe optimizations~\cite{yu2024rlhf,sun2023aligning}.

\input{images/teaser_why_this_task}

Recent works~\cite{li2024llava-onevision,liu2024llava-next,wang2024qwen2} demonstrate that as the scale of Large Language Models (LLMs) increases, so do the capabilities of Multimodal Large Language Models (MLLMs). Furthermore, other studies~\cite{dong2024internlm,hong2024cogagent} reveal that higher image resolution generally enhances performance. However, utilizing larger LLMs or incorporating additional visual tokens imposes a significant computational burden. For instance, as shown in \cref{fig:why_this_task}, a simple query in a visual question-answering task may involve only 51 text tokens but up to 576 visual tokens—over ten times the number of text tokens. This substantial imbalance, combined with the high volume of visual tokens, poses a critical challenge: the computational cost for LLMs rises quadratically with the total input tokens, thereby limiting the scalability of MLLMs.

To address these challenges, various approaches~\cite{chu2023mobilevlm,zhu2024llava-small,zhou2024tinyllava,chu2024mobilevlm,lin2024moe-llava,wei2024small,yao2024minicpm} have attempted to use more lightweight LLMs with fewer parameters, although these often experience noticeable performance drops on general tasks. Recognizing the substantial redundancy within visual representations, other methods~\cite{shang2024prumerge,li2024tokenpacker,li2024mini-gemini,xu2024llava-uhd,yu2024texthawk,zhang2024tinychart} aim to compress visual tokens to reduce input length for LLMs. However, determining which visual tokens are essential within the model remains an open question, and this compression frequently compromises model performance. Additionally, with the inclusion of video inputs, further reduction of image tokens becomes increasingly challenging.

In our work, we  seek to fully leverage visual redundancy to enhance the efficiency of MLLMs.  Rather than pruning input visual tokens to reduce redundant visual representations, we explore redundancy within visual computation patterns in MLLMs. Using LLaVA as a case study, we conduct a detailed analysis of redundancies in both visual attention and representation computations across all layers. Based on these insights, we introduce a set of simple yet effective strategies to significantly reduce the computational burden from visual information: neighbor-aware visual attention, non-active visual attention head dropping, sparse projection in the FFN, and lazy layer dropping for visual processing. As shown in \cref{fig:why_this_task}, most parameters are dropped during visual computation while remaining active in text computation, thereby maintaining the strong understanding capability of the LLM while efficiently reducing the computational overhead introduced by the large number of visual tokens.   By integrating these strategies into LLaVA, we reduce visual computational overhead by up to $88\%$ while retaining nearly the same performance as the original model. Across multiple benchmarks, our method achieves state-of-the-art performance compared to various baselines. Furthermore, these findings are not unique to the LLaVA-based model; similar visual computational redundancies are also present in other MLLMs, including Qwen2-VL-7B and InternVL-2.0-4B/8B/26B. \textcolor{black}{While token pruning-based methods require dynamic identification of important visual tokens on a per-request basis, our method directly prunes the computation pattern in MLLM from the outset of model development. This means that You Only Need to Prune Once (YOPO), which effectively reduces visual computation redundancy and accelerates MLLM performance across all subsequent use cases.} These results highlight a promising direction for future research to enhance MLLM efficiency.

We summarize our contributions as follows:
\begin{enumerate}
    \item We systematically study redundancy in the visual computation pattern of LLaVA, focusing on both intra-modal attention computation among visual tokens and cross-modal attention between visual and text tokens.
    \item We introduce three strategies for visual computation pruning: neighbor-aware visual attention, non-active visual attention head dropping, sparse visual projection in the FFN, and lazy layer dropping for visual processing. These strategies effectively reduce computational costs at the inherent computation pattern level, alleviating the increasing visual computation overhead associated with a large number of visual tokens. \textcolor{black}{Notably, our method prunes the MLLM only once, and it can be deployed for all subsequent computation requests.} 
    \item In experiments, we apply these strategies to prune LLaVA, achieving state-of-the-art results on various key benchmarks with an $88\%$ reduction in computational overhead across vision benchmarks while retaining most model performance. We further validate visual computation redundancy in other MLLMs, including Qwen2-VL-7B and InternVL-2.0-4B/8B/26B, showing that $25-50\%$ of visual computations from parameters and patterns can be pruned without fine-tuning while maintaining comparable performance.
\end{enumerate}

\begin{figure*}
    \centering
    \includegraphics[width=\linewidth]{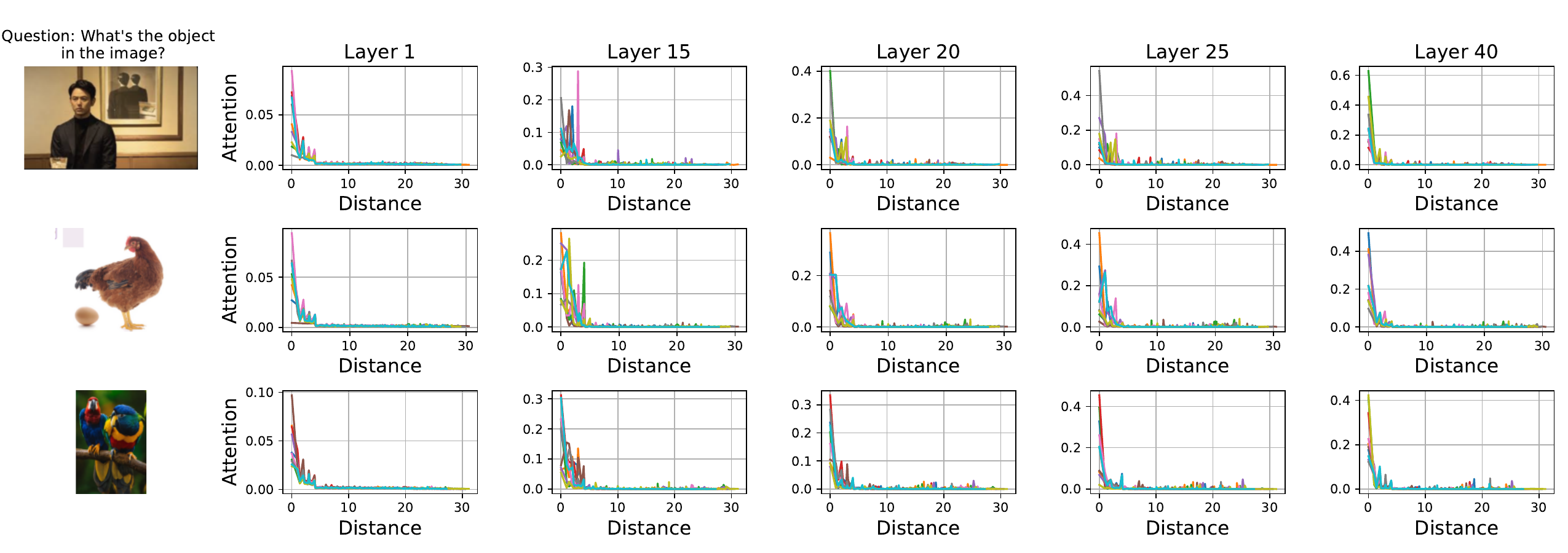}
    \caption{Visualization of attention weights for randomly selected vision tokens interacting with other visual tokens at varying spatial distances across different layers in LLaVA. Notably, the attention weights are predominantly concentrated on neighboring visual tokens. }
    \label{fig:viz_attn}
\end{figure*}
\begin{figure}
    \centering
    \includegraphics[width=0.8\linewidth]{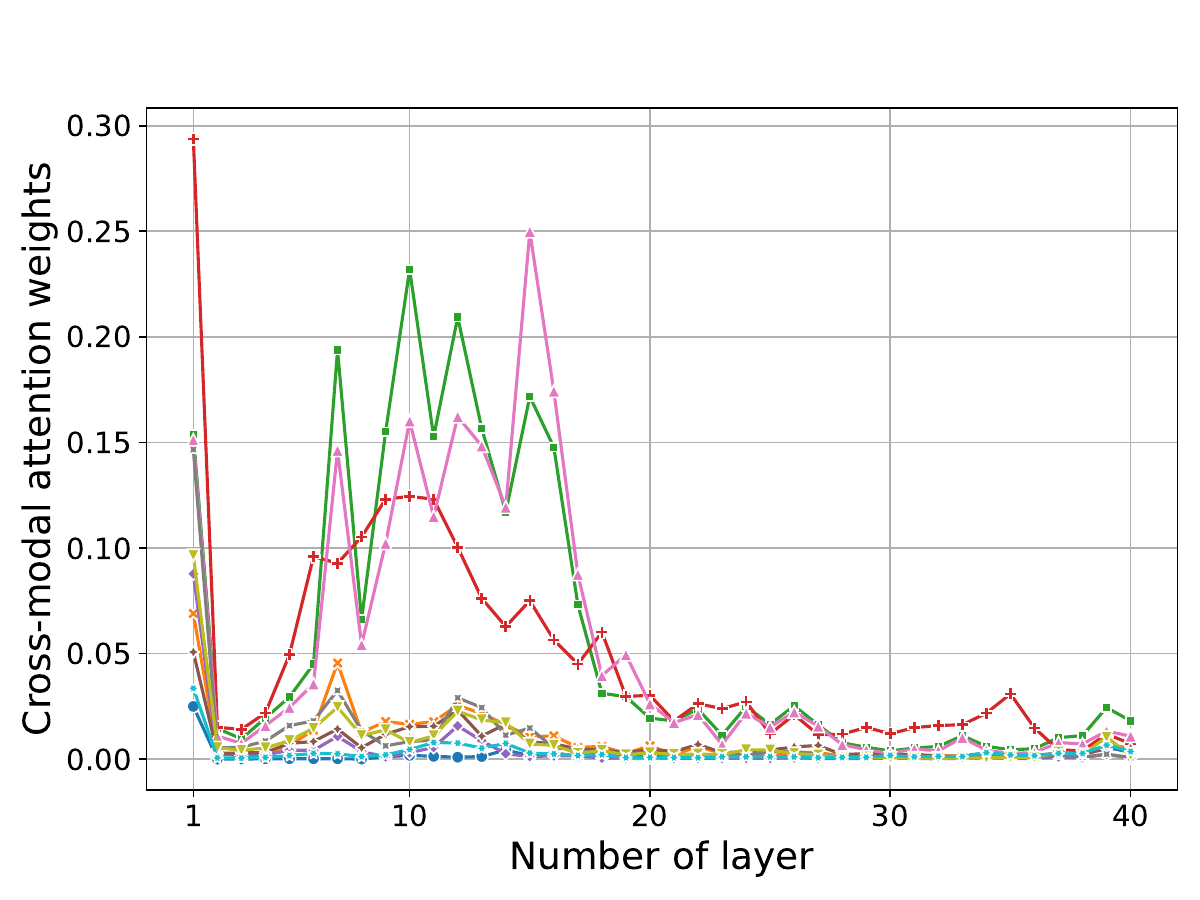}
    \caption{Visualization of the cross-modal attention weights  between vision and text across varying layers. Each line represents the mean attention of an individual text token directed toward all other vision tokens, illustrating how attention varies as layers progress. }
    \label{fig:viz_x_attn}
\end{figure}

\section{Related work}
\subsection{Efficient MLLMs}

Various approaches have been proposed to accelerate MLLMs, including designing smaller vision encoders~\citep{dosovitskiy2020image, fang2023eva, liu2024llavanext}, transforming the dense LLMs into the mixture-of-experts (MoE)~\citep{zhang2024diversifying}, adopting compact language models~\citep{he2024efficient, javaheripi2023phi, chu2023mobilevlm}, and implementing visual token pruning~\citep{li2024mini, xu2024llava, yu2024texthawk}, among others. Among these, token pruning-based methods reduce the length of the input visual sequence without affecting model parameters or capacity, making them increasingly popular.

For instance, LLaVA-PruMerge~\citep{shang2024llava} identifies and merges similar tokens at CLIP's penultimate layer, while PyramidDrop~\citep{xing2024pyramiddrop} applies a predefined token-dropping ratio across layers. Similarly, SparseVLM~\citep{zhang2024sparsevlm} progressively prunes visual tokens deemed irrelevant to corresponding text tokens, and FastV~\citep{chen2025image} enhances the computational efficiency of multi-modal LLMs by using adaptive attention patterns in initial layers to identify essential visual tokens, pruning less important ones in later layers. Although these approaches mitigate computational load, they often compromise model quality due to the loss of fine-grained visual details. 

In our work, while we also leverage the sparsity of visual computation, we propose that pruning within the LLM itself offers a more efficient alternative. This approach preserves visual detail more effectively than methods that directly reduce the quantity of visual tokens.

\subsection{LLM pruning}
The quadratic complexity of LLMs \cite{vaswani2017attention} has made investigating and mitigating their computational redundancy a significant focus in the research community. Techniques such as pruning based on weight importance have been proposed by ~\citet{han2015deepcompression} and ~\citet{frantar2023sparsegpt}, while ~\citet{michel2019sixteen} demonstrated that many attention heads can be removed to improve efficiency. Similarly, ~\citet{fan2019reducing} introduced a method for randomly dropping transformer layers without impacting model performance. While computational redundancy in LLMs has been well-explored, adapting these methods to MLLMs remains an open challenge. In particular, optimizing LLM pruning techniques for the multimodal nature of MLLMs, especially with a focus on the vision modality, has yet to receive adequate attention. 

Our work addresses this gap for the first time by proposing pruning optimizations specifically designed for the LLM components of MLLMs to enhance their efficiency in processing visual information.

\section{Method}

\begin{figure*}
    \includegraphics[width=\linewidth]{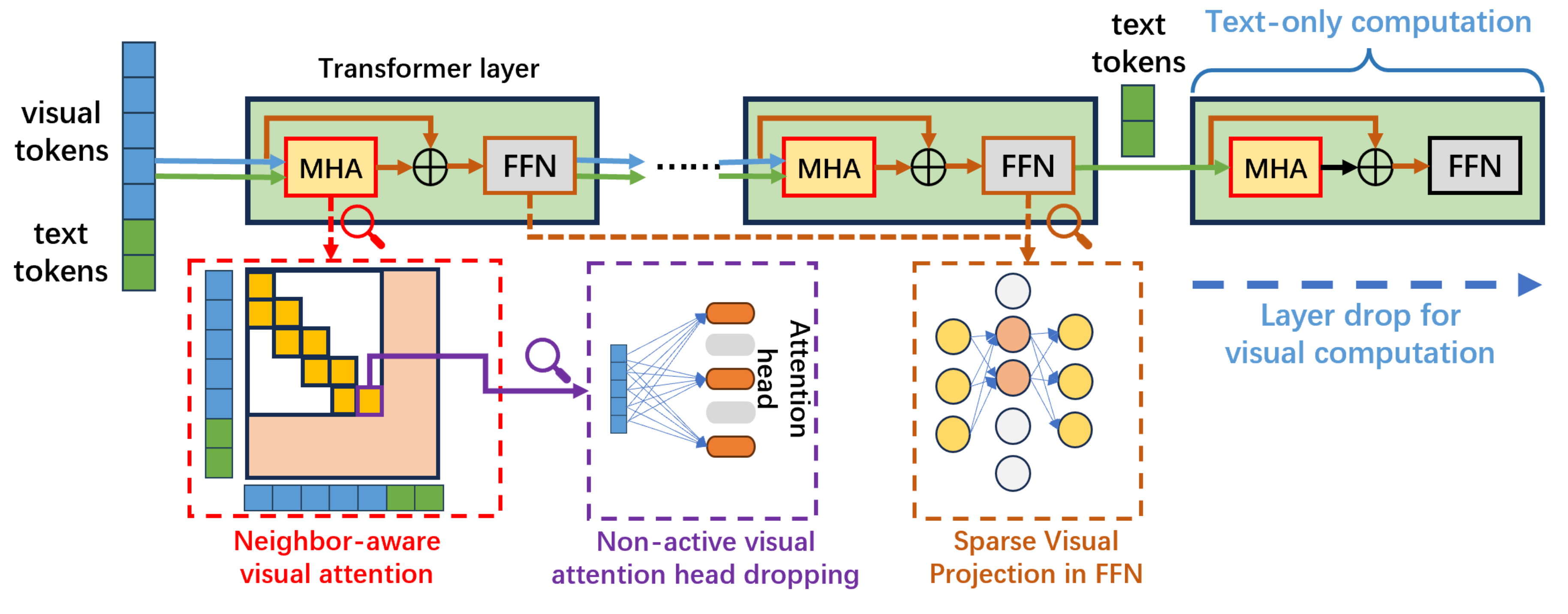}
    \caption{Overview of our method. Our approach replaces the traditional attention block with a neighbor-aware visual attention mechanism, reducing computational complexity from quadratic to linear with respect to the number of visual tokens. Additionally, we prune inactive attention heads in the visual computation, focusing only on the most impactful components. To further decrease computation overhead, we disable visual processing in the later layers, where visual information has minimal impact on the task. }
    \label{fig:prune_method}
\end{figure*}

\subsection{Redundancy of visual computation in MLLMs}
\label{sec:motivation}

In MLLMs, vision tokens are projected into the same latent space as text tokens, with both types processed equally within the network's computational flow. Specifically, given the vision token $V_{v}$ with length $N_v$ and text token $V_{t}$ with length $N_t$, the major computation overhead in MLLMs falls on two modules, including the attention operation
\begin{equation}
    \begin{aligned}
        &\text{Attention}(\mathbf{Q}_{v}, \mathbf{Q}_t, \mathbf{K}_{v}, \mathbf{K}_{t}, \mathbf{V}_{v}, \mathbf{V}_{t}) \\
&= \text{Softmax}\left( \frac{\left[\mathbf{Q}_{v};\mathbf{Q}_{t}\right] \left[ \mathbf{K}_v; \mathbf{K}_{t} \right]^{T}}{\sqrt{d}} \right)  \left[\mathbf{V}_{v}; \mathbf{V}_{t}\right],
    \end{aligned}\label{eq:tra_attn}
\end{equation}
and the feed-forward layer,
\begin{equation}
    \begin{aligned}
        &\text{FFN}(H_v, H_t)=\sigma\left(\sigma\left(\left[H_v; H_t\right]W_{1}\right)W_2\right),
    \end{aligned}\label{eq:ffn}
\end{equation}
where $\mathbf{Q}_v \in \mathbb{R}^{N_v \times d}$ and  $\mathbf{Q}_t \in \mathbb{R}^{N_t \times d}$ are the queries for the vision and text tokens, $\mathbf{K}_v \in \mathbb{R}^{N_v \times d}$ and $\mathbf{K}_t \in \mathbb{R}^{N_t \times d}$ are the keys, $\mathbf{V}_v \in \mathbb{R}^{N_v \times d}$ and $\mathbf{V}_t \in \mathbb{R}^{N_t \times d}$ are the values, $H_v$ and $H_t$ are the visual and text hidden representation, $W_1 \in \mathbb{R}^{d \times d'}$ and $W_2 \in \mathbb{R}^{d' \times d}$ are the weight matrices for linear projection in FFN, $\sigma(\cdot)$ is the activation function.  Given that the number of vision tokens often far exceeds that of text tokens, vision-related computations dominate the workload, causing computational costs to scale approximately quadratically with the number of vision tokens.

Despite this significant computational load, much of the visual processing in MLLMs is redundant due to the inherent spatial sparsity in visual data. To illustrate this,  we randomly select $10$ visual tokens from three example images and visualize their attention weights with respect to other vision tokens in shown images, \textit{i.e.}, $Q_vK_v^T$ in \cref{eq:tra_attn}. More specifically, we first compute the average attention weights across all attention heads and then rearrange these weights for each selected token based on its spatial distance from other tokens. This enables us to investigate the relationship between token spatial distance and the redundancy in visual attention computation. The spatial distance between visual token $i$ and $j$ is defined as follows,
\begin{equation}
    \begin{aligned}
    \text{Distance}(i,j) = \sqrt{(i.x-j.x)^2+(i.y  - j.y)^2},
    \end{aligned}
\end{equation}
where we denote $x$ and $y$ as the coordinates of the tokens in original 2D image.

The results in \cref{fig:viz_attn} show that high attention weights are predominantly concentrated among neighboring tokens, with values diminishing rapidly for tokens farther away. This pattern reflects significant sparsity, suggesting substantial potential for pruning.

This redundancy is observed not only within the vision modality but also in its interactions with text tokens, \textit{i.e.}, $Q_tK_v^T$ in \cref{eq:tra_attn}.  As shown in \cref{fig:viz_x_attn}, we further randomly select $10$ vision tokens and compute the mean attention across other vision tokens at various layers.  The results reveal high sparsity, with attention values generally decreasing to a low level after approximately the $20$-th layer. These findings suggest that visual computation in MLLMs contains significant redundancy, presenting substantial opportunities for pruning to accelerate processing.

\begin{figure}
    \centering
    \includegraphics[width=\linewidth]{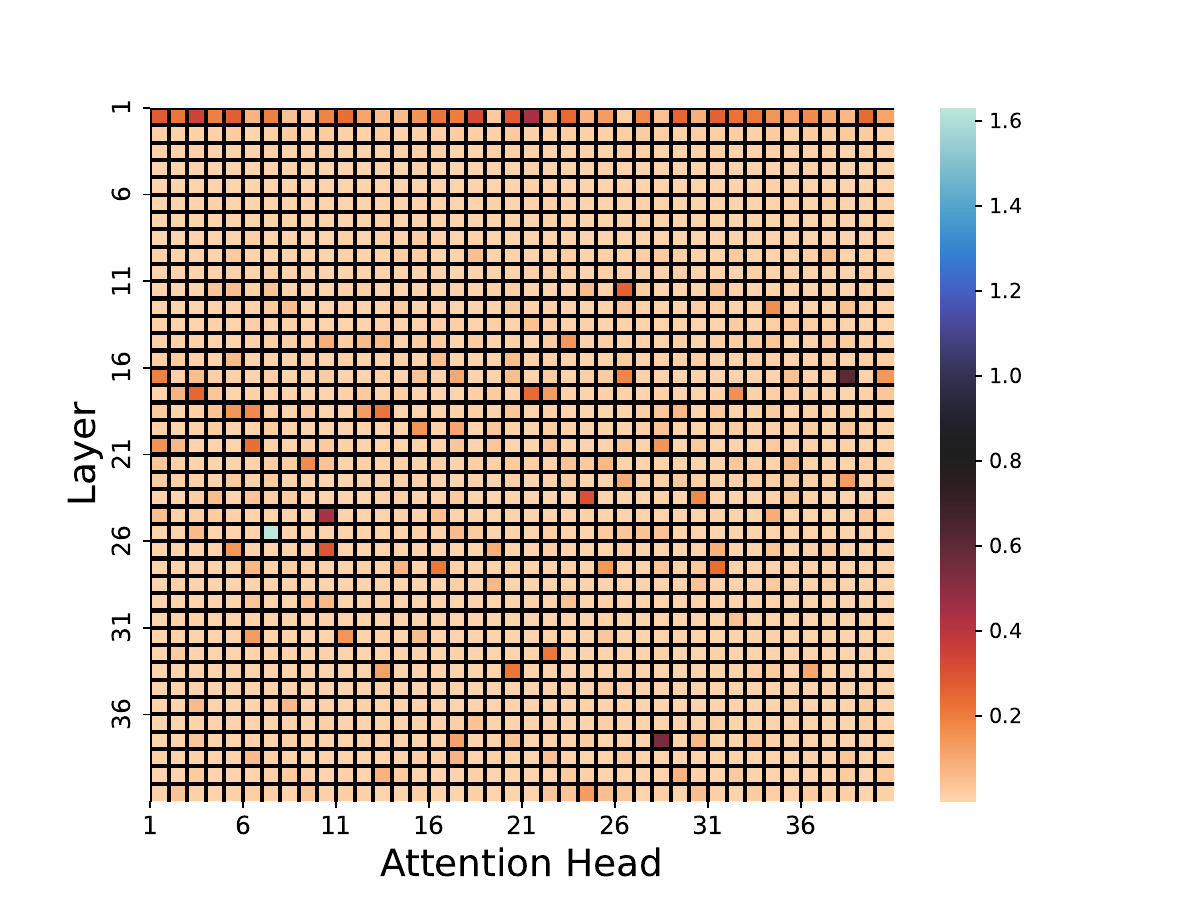}
    \caption{Visualization of $\rho$ of different attention heads in different layers of LLaVA.}
    \label{fig:rho}
\end{figure}

\subsection{Neighbor tokens matter for visual attention}
\label{attention-link-pruning}


As analyzed in \cref{fig:viz_attn}, although there are many visual tokens in MLLMs, most of the attention computation in $Q_vK_v^T$ is sparse, with significant attention weights concentrated primarily on neighboring visual tokens. To reduce the computational burden caused by this redundancy, we propose a simple yet effective pruning method that selectively eliminates non-essential attention computations among visual tokens. Specifically, we modify the attention mechanism so that only neighboring visual tokens attend to each other, while text tokens retain the ability to attend across both visual and text tokens. The modified attention computation for visual tokens can be formulated as follows:
\begin{equation}
    \begin{aligned}
        \text{Attention}(\mathbf{Q}_{v}, \mathbf{K}_{v}, \mathbf{V}_{v}) &= \text{Softmax}\left( \frac{\mathbf{Q}_{v}  \mathbf{K}^T_v}{\sqrt{d}} + \mathbf{M} \right)  \mathbf{V}_{v},\\
        \textit{s.t.}, ~~~~~\mathbf{M}_{ij} &= \begin{cases} 
0 & \text{if } |i - j| \leq h \\
-\infty & \text{if } |i - j| > h 
\end{cases},
    \end{aligned}\label{eq:new_vis_attn}
\end{equation}
where $h$ is the window size indicating the neighborhood of visual tokens. At the same time, the remaining attention computations for the visual-text and text-only tokens are computed as 
\begin{equation} 
\begin{aligned}
&\text{Attention}(\mathbf{Q}_t, \mathbf{K}_{v}, \mathbf{K}_{t}, \mathbf{V}_{v}, \mathbf{V}_{t}) \\
&= \left[\text{Softmax}\left( \frac{\mathbf{Q}_{t} \left[ \mathbf{K}_v; \mathbf{K}_{t} \right]^{T}}{\sqrt{d}} \right)  \left[\mathbf{V}_{v}; \mathbf{V}_{t}\right] \right].
\end{aligned}\label{eq:new_txt_attn}
\end{equation}
The results of \cref{eq:new_vis_attn} and \cref{eq:new_txt_attn} will be contacted as the output attention map with high sparsity.

The neighboring connections $\mathbf{M}$ in \cref{eq:new_vis_attn} can be identified through geometric computation within the CUDA kernel function during the runtime of matrix multiplication, allowing us to dynamically focus only on relevant tokens. This approach enables accelerated attention computation by limiting the number of involved tokens, thereby reducing the overall computational load.  Specifically, when the number of visual tokens $N_V$ is much larger than the number of text tokens $N_T$, the computational complexity reduces from $\mathcal{O}((N_V + N_T)^2)$ to $\mathcal{O}(N_V N_T + N_T^2)$, which grows linearly with $N_V$ instead of quadratically.


\paragraph{Sparse Visual Projection in FFN} By disabling most of the visual attention computation, the model's visual representation becomes highly sparse. To leverage this sparsity effectively, we also randomly drop $p\%$ of neurons in the hidden layer of the FFN within each block, \textit{i.e.}, reducing the dimension $d'$ used in $W_1$ and $W_2$ of \cref{eq:ffn}.

\subsection{Not all visual attention heads are equal}
While we simplify visual attention by focusing only on neighboring visual tokens, the overall computation overhead for visual information remains substantial, largely due to multi-head attention.  A natural question arises, \textit{is every visual attention head equal in visual computation}?

To address this question, we randomly select 100 samples from a visual-question answering task and forward them through LLaVA, recording the attention head values across different layers. To better quantify the impact of visual information on the task, we calculate the relative attention head value which is computed by the mean attention values for all visual tokens and all text tokens, respectively, which can be formulated as follows,
\begin{equation}
    \begin{aligned}
        \rho^h = \frac{\frac{1}{N_v}\sum_{t=1}^{N_v}A^h_{v}}{\frac{1}{N_t}\sum_{t=1}^{N_t}A^h_{t}},
    \end{aligned}\label{eq:relative_attn_head}
\end{equation}
where we respectively denote $A^h_{t}$ and $A^h_{v}$ as the text and vision output of the $h$-th attention head.  A higher value of $\rho^h$ indicating greater engagement of $h$-th attention head in vision question answering.

As shown in \cref{fig:rho}, we surprisingly find that most of attention heads don't answer actively, except the first layer. Thus, to further prune the visual computation in MLLMs, we propose to drop the attention heads of which $\rho<\alpha$, where $\alpha$ is the threshold to decide whether the attention head is active in visual processing.

\subsection{Your model only needs text for deeper layers}

As depicted in \cref{fig:viz_x_attn}, compared with the first several layers, the cross-modal attention weights, \textit{i.e.},  $Q_tK_v^T$, are significantly reduced in deeper layers,  indicating a diminished influence of visual information on the text. 

To further support this observation, we experimented with the LLaVA 1.5-13B model by selectively disabling visual computation in different sets of consecutive transformer blocks, each with a fixed length of 5 layers. For example, we skipped the computation of attention weights between visual and text tokens from layers 5 to 10 and evaluated the pruned model without fine-tuning on various visual question-answering benchmarks in a zero-shot setting.  As reported in \cref{tab:zero_shot_drop}, the performance drop on different benchmarks decreases as we increase the depth of the pruned blocks. Notably, the model retains nearly the same performance when we skip cross-modal computation between layers 20 and 40, which is also consistent with \cref{fig:viz_x_attn}.

\begin{table}
\caption{Evaluation results of dropping various ranges of transformer blocks for visual computation in LLaVA-1.5-13B. The model is not fine-tuned following visual computation pruning.}\label{tab:zero_shot_drop}
\centering
\resizebox{0.8\linewidth}{!}{
    \begin{tabular}{c|ccccc}
    \toprule
        Range &  GQA & SQA$^I$ & TextVQA & MMB & Avg\\\hline
        \textcolor{mygray}{Ori.}   &  \textcolor{mygray}{63.0}    &   \textcolor{mygray}{72.4}  &  \textcolor{mygray}{59.8}       &  \textcolor{mygray}{68.6}   & \textcolor{mygray}{66.0}   \\\hdashline
        0--5  &  38.3    &   63.8  &  38.7       &  56.5   &    57.8 \\
        10--15 &    58.1   &   71.2  &   55.7      &  62.7   &   65.2 \\
        20--25&   62.1    &  72.3   &   59.1      &  68.6   &    66.0\\
        30--35&   63.1    &  72.4   &    59.8     &  68.6   &   66.0 \\
        35--40&   63.0    &   72.2  &    59.3     &   68.6  &   66.0 \\\bottomrule
    \end{tabular}}
\end{table}

Given that the output of MLLMs is ultimately text, this observation motivates us to directly skip all visual related computation in later layers, thereby reducing computational overhead. Specifically,  for layers $l > L-N$, we propose to omit visual-related computations, including both visual and cross-modal attention, allowing the attention computation to be simplified as follows:
\begin{equation} 
\begin{aligned}
\text{Attention}(\mathbf{Q}_t, \mathbf{K}_{t}, \mathbf{V}_{t})
= \text{Softmax}\left( \frac{\mathbf{Q}_{t}  \mathbf{K}_{t}^{T}}{\sqrt{d}} \right)  \mathbf{V}_{t},
\end{aligned}\label{eq:txt_attn}
\end{equation}

With this design, only the text represented by $\mathbf{H}_T^{(l)}$ are processed independently in the last $N$ layers.


\section{Evaluations}

\subsection{Experiment setup}

\noindent \textbf{Studied models}.  In this work, we mainly focus on pruning visual computation in LLaVA models~\citep{llava-1.0}, specifically the {LLaVA-1.5-7B} and {LLaVA-1.5-13B} models. The hidden dimensions for LLaVA-1.5-7B and LLaVA-1.5-13B are 4096 and 5120, respectively, with intermediate sizes in the FFN of 11008 and 13824. The LLaVA-1.5-7B model has 32 layers, each containing 32 attention heads, while the LLaVA-1.5-13B model has 40 layers, each with 40 attention heads. Both models share a vocabulary size of 32,000.   For efficiency comparison, we also include MoE-LLaVA-1.6Bx4 and MoE-LLaVA-2.7B~\citep{lin2024moe}, of which backbones are the StableLM-1.6B~\citep{bellagente2024stable} and Phi-2.7B~\citep{javaheripi2023phi}, respectively. 

Simultaneously, we apply our proposed pruning strategies to Qwen2-VL-7B~\citep{bai2023qwen} and InternVL-2.0~\citep{chen2024internvl} models to further demonstrate the pervasive nature of visual computation redundancy in current MLLMs.

\noindent \textbf{Selected baselines}. We select four methods as our baselines, including PruMerge+~\citep{shang2024llava}, PyramidDrop~\citep{xing2024pyramiddrop}, SparseVLM~\citep{xing2024pyramiddrop}, and FastV~\citep{chen2025image}. Among these baselines, PruMerge+ and PyramidDrop require a fine-tuning step to accommodate the reduction of visual tokens, while SparseVLM and FastV are training-free methods designed to accelerate the model.

\noindent \textbf{Evaluation benchmarks}. We evaluate our method on diverse benchmarks covering visual question answering, scientific reasoning, and multimodal understanding. VQAv2 \cite{vqav2} tests visual and commonsense reasoning with open-ended questions on images. ScienceQA \cite{scienceQA} includes multi-modal questions on science topics, and we focus on the SQA$^I$ subset, which contains questions that specifically include images as part of the question context. TextVQA \cite{textVQA} challenges models with  text-recognition questions on different images. GQA \cite{GQA} tests fine-grained visual reasoning with multistep question-answer pairs. POPE \cite{POPE} measures object hallucination under varying conditions. MME \cite{MME} and MMBench \cite{MMBench} assess multimodal reasoning across thousands of image-text pairs. We show the number of examples in each selected benchmark for model evaluation in \cref{tab:main}. These benchmarks offer a comprehensive evaluation of our method in diverse task domains.

\noindent \textbf{Implementations}. In our experiments, we apply individual and combined pruning strategies to reduce visual computation redundancy in the model. We evaluate our approach in two settings: training-based and training-free. In the training-based setting, we use the same data and training protocol as the public repository of LLaVA-1.5 models, training from scratch with reduced visual computation redundancy. The number of iterations remains the same to allow a fair comparison with the original model. In the training-free setting, we directly apply different pruning strategies to studied models, \textit{i.e.}, Qwen2-VL-7B, and InternVL-2.0,  without any fine-tuning. All experiments were conducted on a single node with 8 A100-80G GPUs,  with detailed results provided in \cref{append:exp_de}.


\subsection{Comparison with efficient LLaVA models}
We begin by evaluating the effectiveness of our proposed method for pruning the LLaVA-1.5-7B and LLaVA-1.5-13B models. By combining our four strategies, we reduce the FLOPs to $25\%$ and $12\%$ of the original LLaVA model, respectively. The results are reported in \cref{tab:main}.

With the same computational budget, \textit{i.e.}, the same FLOPs, our pruning method consistently achieves the best results on the four  benchmarks, showing an average performance gap of $3.7\%$, $1.1\%$, $2.2\%$, $0.45\%$ over the runner-up method on GQA,  VQAv2, POPE, and MMB, respectively. Although our method does not perform best on certain test cases, such as LLavA-7B on SQA$^{I}$ and TextVQA, our pruned model achieves comparable results, with only a minor average performance drop of $0.5\%$. 

It should be noted that our pruned model, using only $12\%$ of parameters for visual computation, can achieve performance comparable to selected baselines with a computation budget approximately $2\times$ larger. Specifically, on the largest dataset, GQA, our 7B model surpasses the runner-up method with a clear gap of $3.4\%$. 

Furthermore, when comparing the results with MoE-LLAVA-1.6Bx4 and MoE-LLaVA-2.7Bx4—models with a total number of parameters similar to LLaVA-7B and LLaVA-13B, respectively—our pruned model demonstrates greater efficiency and superior performance on selected benchmarks. While MoE is often utilized to scale up model parameters while improving inference efficiency, given the challenges of MoE training, this comparison suggests that pruning from a large dense model is also a promising, practical, and simple approach to retain most of the performance while significantly reducing the computation budget, making it more suitable for efficient deployment.

\begin{table*}[ht]
\centering
\caption{Performance comparison with various efficient MLLMs on different benchmarks. The `-` symbol indicates missing values where certain evaluations were not available. \textbf{We list the number of test examples in each benchmark.} }\label{tab:main}
\resizebox{0.9\textwidth}{!}{
\begin{tabular}{ccccccccccc}
\hline
\multirow{2}{*}{Backbone} & \multirow{2}{*}{Model}   & \multirow{2}{*}{FLOPs (T)} & {GQA} & {VQAv2} & {SQA$^I$} & {TextVQA} & {POPE} & {MME} & {MMB$_{en}$} \\ 
&  & & 12K & 107K & 4K & 5K & 9K & 2K & 4K\\ \hline
StableLM & MoE-LLaVA-1.6Bx4 & 1.96 & 60.4 & -- & 62.6 & 47.8 & 84.3  & 1300.8 & 59.4  \\
Phi & MoE-LLaVA-2.7Bx4 &  5.42 &  61.1 & -- & 68.7   & 50.2 &  85.0&  1396.4 & 65.5 \\ \hline
\multirow{7}{*}{LLaVA-1.5-7B} &
Original  & 7.63 & \textcolor{gray}{62.3} & \textcolor{gray}{79.2} & \textcolor{gray}{69.4} & \textcolor{gray}{58.1} & \textcolor{gray}{87.3}  & \textcolor{gray}{1492.8} & \textcolor{gray}{65.4} \\
&PruMerge+~\citep{shang2024llava}   &   1.91 &  -- & 76.8 & 68.3   & \bf{57.1} &  84.0 &  1462.4 & 64.9 \\
&Pyramid-Drop~\citep{xing2024pyramiddrop}   &  1.91 & 57.3 & 75.1 & \bf{69.3} & 55.9 & 83.4 &  1300.8 & 59.4 \\
&SparseVLM~\citep{zhang2024sparsevlm}  &  1.91 & 50.2  & 62.9 &  67.1 &  52.5 &  72.0 &  1090.8& 60.0 \\
&FastV~\citep{chen2025image}    &  1.91 & 56.5  & 73.5 &  69.1 &  57.5 &  77.8 &  1380.2& 62.9 \\\cdashline{2-9}
&\multirow{2}{*}{\textbf{Ours}}  &  1.91 & \bf{61.6} & \bf{78.0} & 69.0 & {56.3} & \bf{86.8}  & \bf{1483.6} & \bf{65.5} \\
&  &  0.92 & 60.7 & 77.4 & 68.0 & 55.2 & 86.6  & 1429.6 & 64.1 \\ \hline
\multirow{7}{*}{LLaVA-1.5-13B} & Original  &  14.89 &  \textcolor{gray}{63.0}  &  \textcolor{gray}{79.7}  &  \textcolor{gray}{72.4} & \textcolor{gray}{59.8} & \textcolor{gray}{86.4} &  \textcolor{gray}{1558.0} &  \textcolor{gray}{68.7}  \\
&PruMerge+~\citep{shang2024llava}   & 3.72 & -- & 77.8 & 71.0 & \bf{58.6} & 84.4 & 1485.0 & 65.7 \\
&Pyramid-Drop~\citep{xing2024pyramiddrop}   & 3.72 & 58.4 &  76.7&  \textbf{72.3} & 58.1 &   83.3 & \bf{1506.6} & 65.8 \\
&SparseVLM~\citep{zhang2024sparsevlm}  & 3.72 & 47.2 &  57.3 &  -- &  53.5 & 65.5  & 1090.8  & 63.1 \\
&FastV~\citep{chen2025image}    & 3.72 & 59.4 & 77.1 & 71.2 & 58.5 & 82.0 & 1506.8 & 67.1 \\\cdashline{2-9}
&\multirow{2}{*}{\textbf{Ours}}  &  3.72&  \bf{62.5} & \bf{78.9} & 71.3 & 58.3 & \bf{86.3} & 1457.3 & \bf{67.4} \\
&  &  1.79&  61.2 & 78.1 & 70.7 & 56.7 & 86.4 & 1426.2 & 65.5 \\ \hline
\end{tabular}
}
\end{table*}

\subsection{Pruning with different granularity}
\begin{figure}
    \centering
    \includegraphics[width=\linewidth]{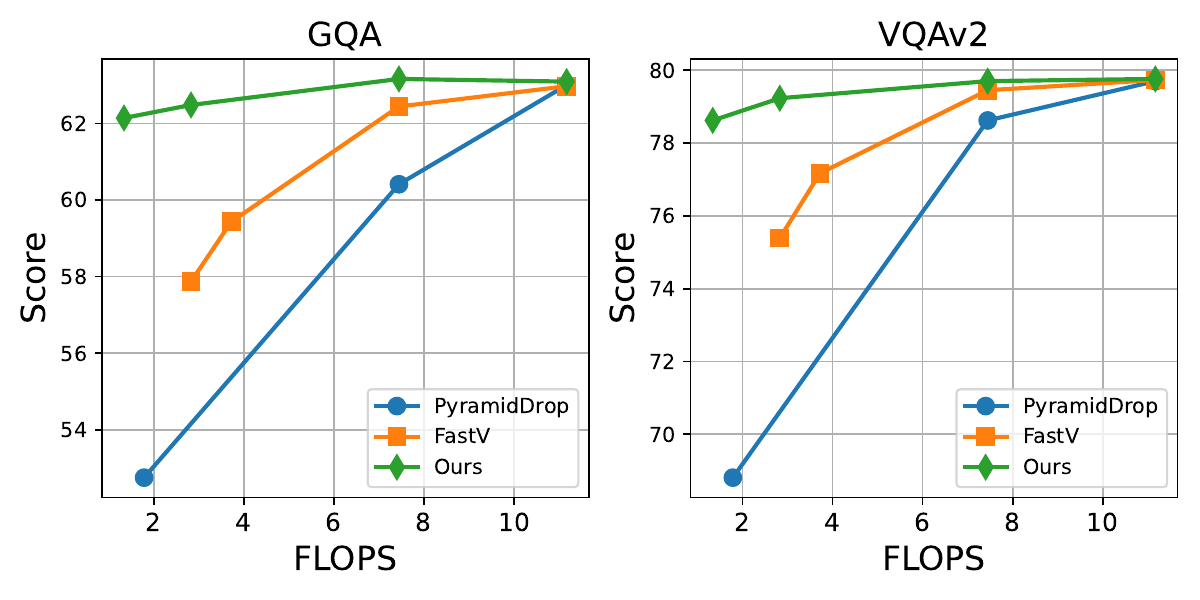}
    \caption{Evaluation results of the pruned LLaVA-1.5-13B   using various strategies with differing pruning granularities. Results are reported on the two largest benchmarks:  GQA and VQAv2.}
    \label{fig:pru_gra}
\end{figure}

To demonstrate the scalability of our method in pruning visual computation redundancy, we compare our proposed strategy with PyramidDrop and FastV at different pruning granularities on the two largest benchmarks, VQAv2 and GQA. Scores on these benchmarks are reported along with corresponding FLOPs, as shown in \cref{fig:pru_gra}.

It can be observed that as FLOPs for visual computation decrease, the performance of the pruned model also declines. Specifically, reducing FLOPs from $75\%$ to $19\%$ led to a performance drop from $71.35\%$ to $66.63\%$ on average across the two benchmarks for the model pruned using FastV. In contrast, rather than pruning tokens, our approach targets redundant visual computations at the parameter and computation pattern levels, resulting in only a $0.5\%$ performance decrease. These results further support our claim that a substantial amount of visual computation redundancy can be effectively pruned in current MLLMs.

\subsection{Computational redundancy beyond LLaVA}
\begin{figure}
    \centering
    \includegraphics[width=0.8\linewidth]{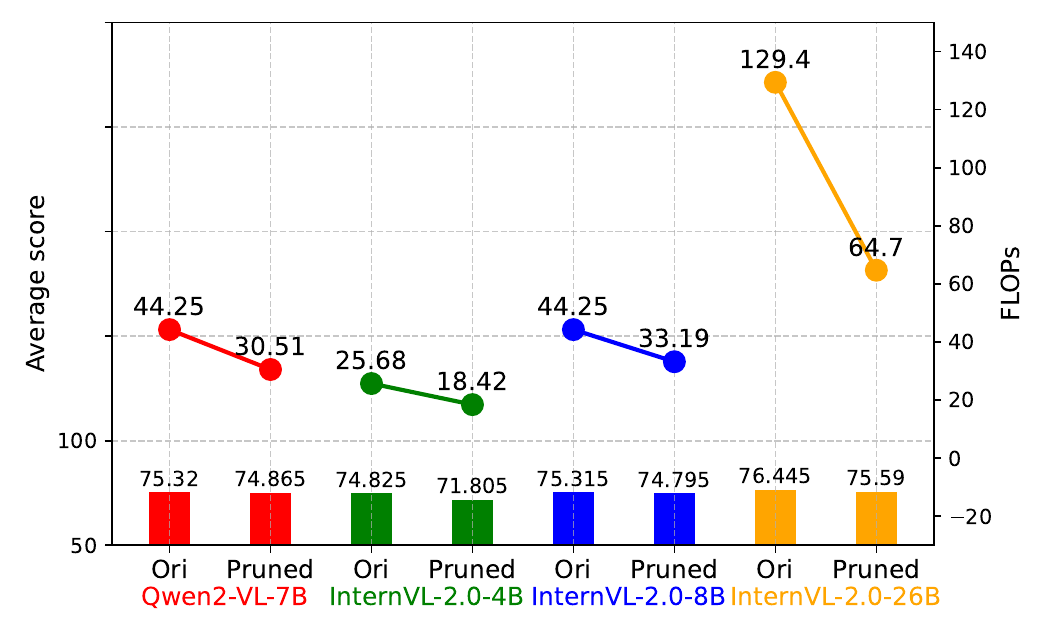}
    \caption{Evaluation results of the pruned Qwen2-VL-7B, InernVL-2.0-4B/8B/26B. The results are averaging on GQA and POPE.}
    \label{fig:more_mllms}
\end{figure}

The visual computational redundancy is not unique to LLaVA. To validate the broader applicability of our pruning strategy, we applied it to additional MLLMs, including Qwen2-VL-7B and InternVL-2.0, without fine-tuning due to limited training data. We evaluated performance on GQA and POPE benchmarks, adjusting pruning granularity to match original model performance with minimal FLOPs.  As shown in \cref{fig:more_mllms}, we observe that, even without fine-tuning, pruning visual computations in these models at suitable ratios does not compromise their performance. Furthermore, larger MLLMs appear to accommodate higher pruning ratios, as evidenced by the results from pruning InternVL-2.0 at various model scales.    More details can be found in \cref{append:more_models}.

\subsection{Discussion}
\noindent \textbf{On the neighbor-aware visual attention}. We vary the number of neighboring visual tokens in the attention computation by setting the radius from $1$ to $13$ in increments of $2$, and report the results in \cref{tab:ne_attn}. The peak performance occurs at a radius of $7$, which aligns with the observation in \cref{fig:viz_attn}, where most non-sparse attention values fall within this spatial range. This suggests that carefully selecting the optimal radius for neighboring token attention computation can accelerate both the training and inference of our model.

\noindent \textbf{On the non-active visual attention head dropping}. In our method, we use the ratio between visual attention and text attention to identify non-active attention heads for visual computation. To further study the effect of dropping attention heads on model performance, we vary the number of randomly selected attention heads dropped for visual computation in each layer from $5$ to $35$, and report the results in \cref{tab:ne_head}. The results show that dropping attention heads within this range mostly preserves the model’s original performance. Generally, as the number of dropped attention heads increases, model performance gradually declines. Notably, even with $35$ attention heads dropped, there is only a $2.0\%$ average performance drop, indicating significant redundancy in visual computation within the model.

\noindent \textbf{On the sparse visual projection in FFN}. Sparse projection is employed to better leverage the sparsity of the visual representation after significantly pruning the model’s visual computation. As shown in \cref{tab:main}, we increase the ratio of dropped neurons in the FFN from $25\%$ to $50\%$, thereby reducing the FLOPs from $1.91$ to $0.92$ on LLaVA-7B, and from $3.72$ to $1.79$ on LLaVA-13B. Without applying sparse projection in the FFN and using the same pruning strategy as in \cref{tab:main}, the LLaVA-13B model achieves a performance of $62.5$ on GQA and $79.2$ on VQAv2, which is quite similar to the model with $25\%$ dropped neurons in the FFN. These results suggest, on one hand, the redundancy in visual computation within the FFN. On the other hand, we observe that although increasing the drop ratio results in further performance decline, it still outperforms models with similar FLOPs, \textit{i.e.}, the MoE-LLaVA models.

\noindent \textbf{On the layer-dropping for visual computation}. Layer dropping is the most effective strategy to reduce FLOPs, where only text is processed within the dropped layers. While we previously provided results for block-wise layer dropping, here we conduct further experiments by continuously dropping layers for visual computation. As shown in \cref{tab:ne_ld}, even when the last $20$ layers are dropped, the model’s performance remains almost unchanged compared to the original model. This strongly suggests that the visual computation is extremely redundant in deep layers, leaving great potential to prune for improving the efficiency.  


\begin{table}[]
\centering
\caption{Results of applying the neighbor-aware attention mechanism as a visual pruning strategy to the LLaVA-1.5-13B model. The 'radius' parameter defines the scope of neighboring tokens considered in visual token computation.  }\label{tab:ne_attn}
\resizebox{0.8\columnwidth}{!}{%
\begin{tabular}{cccccc}
\toprule
Radius & {GQA} & {VQAv2} & {SQA} & {TextVQA} & Avg \\\hline
  {1} & 62.8 & 79.1 & 71.2 & 58.5& 67.9\\
3     & 62.7 & 79.2  & 71.5 & 58.5 & 68.0  \\
5     & 62.8 & 79.7  & 70.9 & 59.9 & 68.3  \\
7     & 63.2 & 79.8   & 71.7 & 59.6 & 68.6 \\
9     & 63.2 & 79.7& 71.3& 59.8& 68.5\\
11    & 63.0 & 79.6 & 70.2 & 59.5 & 68.1 \\
13    & 62.8 & 79.5 & 71.6 & 59.2 & 68.3 \\\bottomrule
\end{tabular}%
}
\end{table}

\begin{table}[]
\centering
\caption{Results of dropping non-active attention heads as a visual pruning strategy to prune the LLaVA-1.5-13B model. '\# Heads' indicates the number of dropped non-active attention heads per layer used in visual token computation.  }\label{tab:ne_head}
\resizebox{0.8\columnwidth}{!}{
\begin{tabular}{cccccc}
\toprule
\# Heads & {GQA} & {VQAv2} & {SQA} & {TextVQA} & {Avg} \\\hline
5  & 62.7 & 79.7  & 71.3 & 59.6 & 68.3 \\
10 & 62.5 & 79.4  & 72.0 & 59.9 & 68.4 \\
15 & 62.5 & 79.1 & 71.6 & 58.1 & 67.9 \\
20 & 62.0 & 78.9 & 72.0 & 58.4 & 67.8 \\
25 & 62.1 & 78.9 & 70.0   & 58.6 & 67.4 \\
30 & 62.0 & 78.5 & 71.2 & 57.5 & 67.3 \\
35 & 61.0 & 77.7 & 71.1 & 55.2 & 66.3  \\\bottomrule
\end{tabular}}
\end{table}

\begin{table}[]
\centering
\caption{Results of applying the layer-dropping strategy to prune the LLaVA-1.5-13B model.\# layer indicate the number of last several layers to drop for visual-related computation.  }\label{tab:ne_ld}
\resizebox{0.8\columnwidth}{!}{
\begin{tabular}{cccccc}
\toprule
\# Layer & {GQA} & {VQAv2} & {SQA} & {TextVQA} & Avg \\\hline
5  & 63.1 & 79.8  & 71.9 & 59.1 & 68.5 \\
10 & 63.0 & 79.7 & 72.3 & 59.0 & 68.5  \\
15 & 63.0 & 79.8 & 71.6 & 59.6 & 68.5 \\
20 & 63.1 & 79.7 & 71.1 & 58.1 & 68.1   \\
25 & 62.2 & 79.0 & 71.5 & 55.5 & 67.1   \\
30 & 61.5 & 78.0 & 69.5 & 56.3 & 66.4  \\
35 & 55.0 & 72.4 & 71.6 & 53.6 & 63.2  \\\bottomrule
\end{tabular}}
\end{table}

\noindent \textbf{Why not directly prune the parameters for both the vision and texts}? We address the redundancy in visual token computations, reducing their overhead while preserving text token computations. To explore if text tokens hold similar redundancy, we ran an experiment pruning 20 attention heads for only visual tokens versus both vision and text tokens. Without fine-tuning, pruning just visual tokens resulted in an average performance of $67.1\%$ across VQAv2, GQA, SQA, and TextVQA, while pruning both led to a drastic drop to $4.3\%$. This indicates much higher redundancy in visual computations than in text within current MLLMs.

\begin{figure}
    \centering
    \includegraphics[width=0.8\columnwidth]{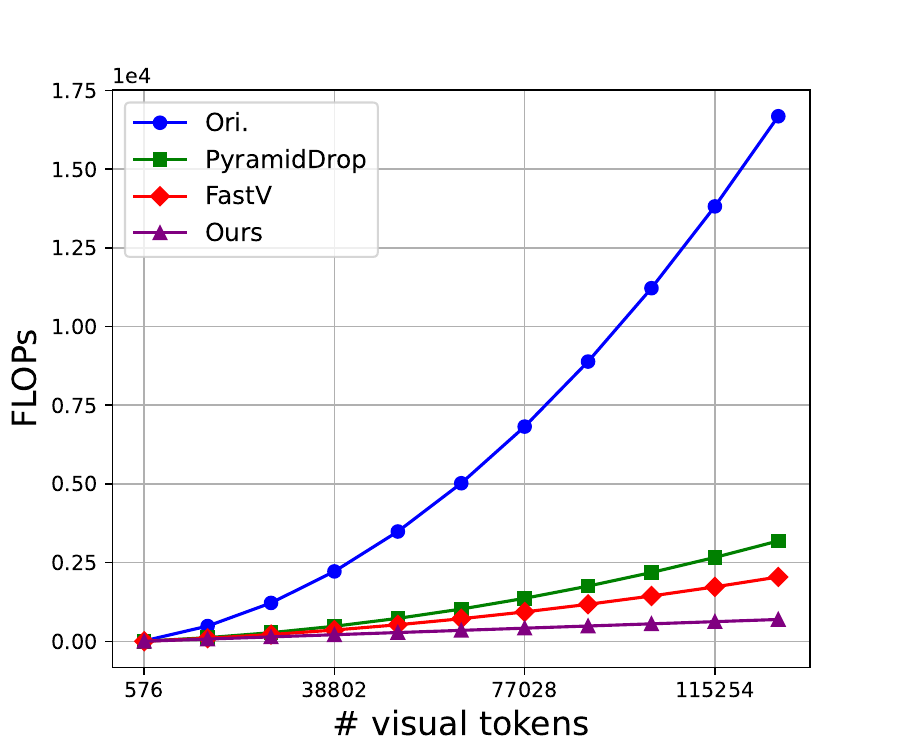}
    \caption{Comparison of the FLOPs for different methods varying the number of visual tokens. The studied model is LLavA-1.5-13B.}
    \label{fig:efficiency_cmp}
\end{figure}

\noindent \textbf{Efficiency analysis on token pruning and computation pattern pruning}. We provide a comparison of efficiency across various methods with differing numbers of input visual tokens in \cref{fig:efficiency_cmp}. The results indicate that, in contrast to token pruning-based approaches, addressing visual computational redundancy at the computational pattern level yields a greater efficiency advantage for long visual sequences. This approach effectively mitigates the escalating computational overhead associated with handling large numbers of visual tokens, demonstrating its superior scalability in processing extended visual sequences.

\section{Conclusion}
In this work, we address the challenge of pruning MLLMs for efficient computation. Unlike text, visual information is sparse and  redundant. Prior work has focused on reducing visual tokens; we instead analyze redundancy in parameters and computation patterns. Our strategies—neighbor-aware visual attention, non-active visual heads dropping, sparse visual projection in FFNs, and layer dropping—reduce LLaVA's computational overhead by $88\%$ while largely preserving performance. Additional experiments on Qwen2-VL-7B and InternVL-2.0 further confirm that visual computation redundancy is prevalent across MLLMs.

{
    \small
    \bibliographystyle{ieeenat_fullname}
    \bibliography{refs}
}


\input{appendix}

\end{document}

%% file: images/teaser_why_this_task.tex
\begin{figure}
    \includegraphics[width=\linewidth]{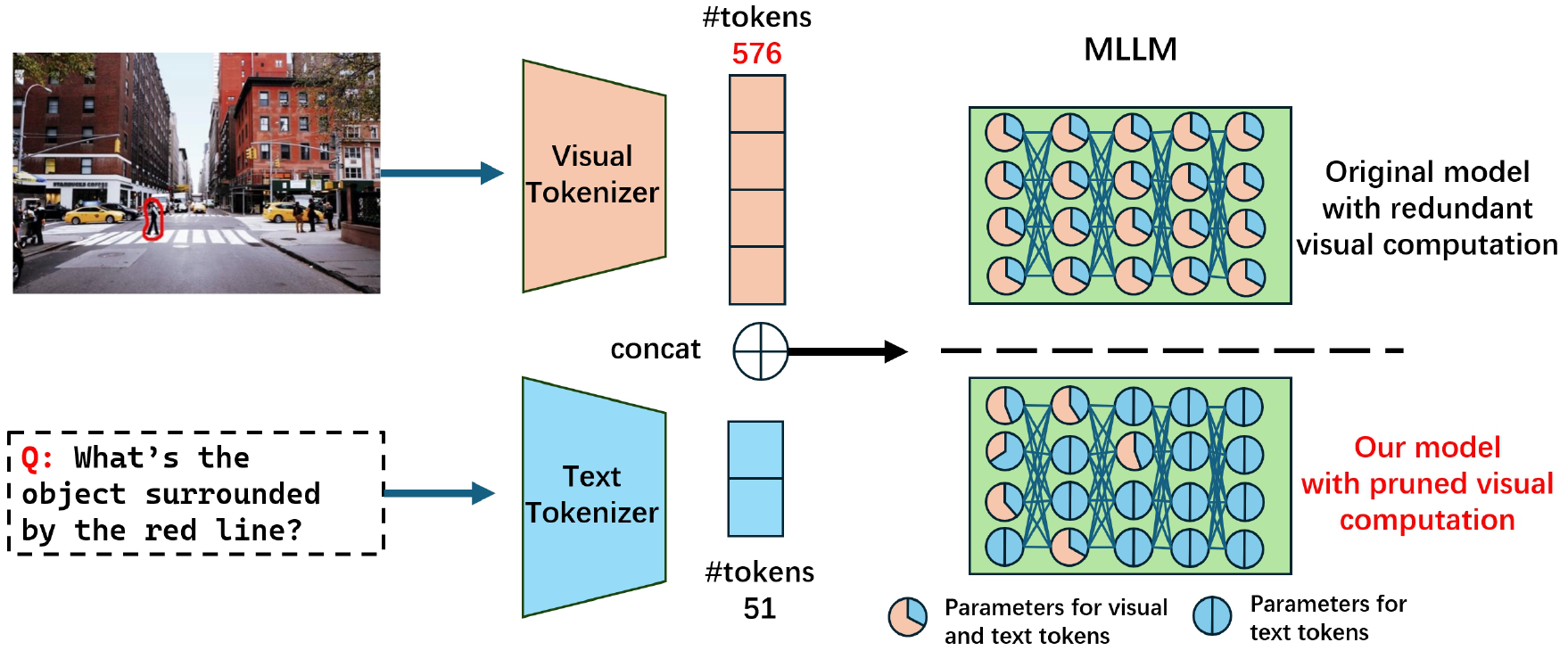}
    \caption{Compared to text tokens processed by language prompts, visual tokens are significantly more numerous, leading to substantial computational overhead in MLLMs. However, visual information is often sparser, resulting in considerable redundancy within visual computations. In this work, we propose pruning these redundant computations at both the parameter and computational pattern levels to improve processing efficiency.}
    \label{fig:why_this_task}
\end{figure}

%% file: appendix.tex
\clearpage
\appendix
\renewcommand{\appendixpagename}{Appendix}
\appendixpage

\section{Experiment details}
\label{append:exp_de}

\noindent \textbf{For the LLaVA pruning with training}. We employed training data and settings identical to those used in LLaVA-1.5. All experiments were conducted on a single node equipped with $8$ A100-80G GPUs. The pre-training phase took $4$ hours, while instruction fine-tuning took 20 hours. 

\noindent \textbf{For the MLLMs pruning without training}. For the experiments of pruning models without the training for performance restoration, we use one A100-80GB GPU.

\noindent \textbf{Pruning details for our models}. In \cref{tab:main}, we present the results of the LLaVA pruned by the combination of our strategies. Here, we present the pruning details in \cref{tab:llava_summary}.

\section{More results on LLaVA-1.5-7B}
\label{append:7b}
We provide a detailed evaluation of pruning LLaVA-1.5-7B using various strategies, including neighbor-aware attention computation (\cref{tab:7Bne_attn}), dropping inactive attention heads (\cref{tab:7Bne_head}), and layer dropping (\cref{tab:7Bne_ld}). The results show that a radius of 5 achieves efficient attention computation with strong performance. Additionally, dropping 12 attention heads and the last 12 layers during visual computation allows for significant acceleration without compromising performance.

\section{More results on Qwen2-VL-8B and InternVL-2.0-4B/8B/26B}
\label{append:more_models}

We present more results of pruning the Qwen-2-VL-8B and InternVL-2.0-4B/8B/26B without the fine-tuning under different pruning strategies and granularity. 

\begin{figure*}
    \centering
    \includegraphics[width=\linewidth]{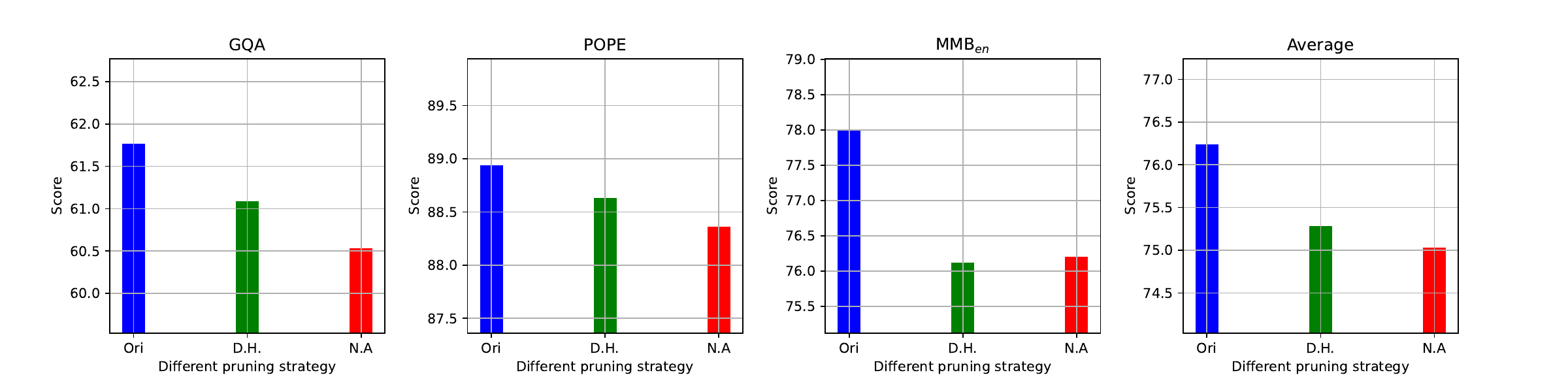}
    \caption{Evaluation results of Qwen2-VL-8B under different pruning strategies. We denote D.H. as dropping non-active visual attention heads during the visual-related computation, and N.A. as the neighbor-aware visual attention computation.}
    \label{supfig:qwen}
\end{figure*}

\begin{figure*}
    \centering
    \includegraphics[width=\linewidth]{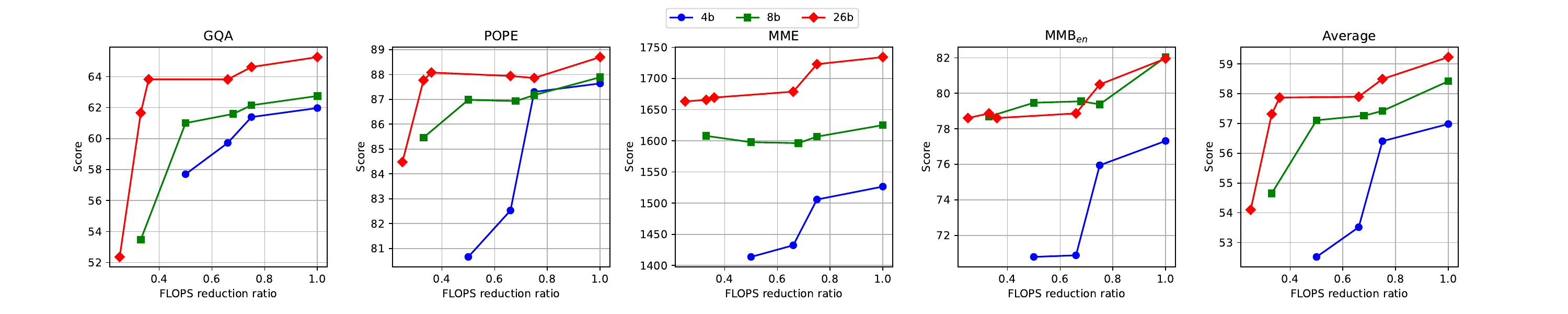}
    \caption{Evaluation results of InternVL-2.0-4B/8B/26B under different pruning ratios. The ratio of $1$ indicates the original model without pruning.}
    \label{supfig:intern}
\end{figure*}

\noindent \textbf{Pruning details.} For Qwen-2-VL-8B, we respectively use the non-active attention head dropping and neighbor-aware attention computation to reduce the FLOPs to 75\%. For InternVL-2.0-4B/8B, we set the radius for attention computation as $4$ to reduce the FLOPs to $75\%$, and integrate the non-active attention head dropping with $24$ attention heads maintaining to further reduce to 66\%, then discard the visual-related computation after $24$ layers to reduce to 50\%, while changing the layer from $24$ to $16$ reduces the FLOPs to 44\%. For InternVL-2.0-8B, we further reduce the last layer for visual computation to $12$ to reduce the FLOPs to 33\%. For InternVL-2.0-26B, by setting the radius for visual attention computation as $4$, maintaining $24$ active visual attention heads, and discarding the visual-related computation after the $36$-th layer, we can reduce 50\% FLOPs, while changing the discarding layer to $24$ and $19$ can further reduce to 33\% and 25\% FLOPs, respectively.  The results are shown in \cref{supfig:qwen} and \cref{supfig:intern}

\begin{table}
\caption{Summary of LLaVA-ours models with corresponding pruning parameters. We denote \# Heads as the number of attention heads retained for the visual attention computation, and Radius as the radius of neighbor visual tokens for attention computation. The \# Layers is denoted as the number of last layers to drop for the visual-related computation. The \# Neurons is denoted as the ratio of neurons retained at the FFN layer. }
\label{tab:llava_summary}
\centering
\resizebox{\columnwidth}{!}{
\begin{tabular}{|l|c|c|c|c|c|}
\hline
\multirow{2}{*}{\textbf{Model}} & \multirow{2}{*}{\textbf{FLOPs}} & \multirow{2}{*}{\textbf{\# Heads}} & \multirow{2}{*}{\textbf{Radius}} & \multirow{2}{*}{\textbf{\# Layers}} & \multirow{2}{*}{\textbf{\# Neurons}} \\ 
                                &                               &                               &                              &                          &                        \\ \hline
\multirow{2}{*}{LLaVA-7B-ours}  & 1.91                          & 24                            & \multirow{2}{*}{5}           & \multirow{2}{*}{16}     & 50\%                   \\ \cline{2-3} \cline{6-6}
                                & 0.92                          & 16                            &                              &                          & 25\%                   \\ \hline
\multirow{2}{*}{LLaVA-13B-ours} & 3.72                          & 30                            & \multirow{2}{*}{5}           & \multirow{2}{*}{20}     & 50\%                   \\ \cline{2-3} \cline{6-6}
                                & 1.79                          & 20                            &                              &                          & 25\%                   \\ \hline
\end{tabular}}
\end{table}

\begin{table}[]
\centering
\caption{Results of applying the neighbor-aware attention mechanism as a visual pruning strategy to the LLaVA-1.5-7B model. The 'radius' parameter defines the scope of neighboring tokens considered in visual token computation.  }\label{tab:7Bne_attn}
\resizebox{0.8\columnwidth}{!}{%
\begin{tabular}{cccccc}
\toprule
Radius & {GQA} & {VQAv2} & {SQA} & {TextVQA} & Avg \\\hline
1  & 61.8 & 78.6 & 69.9 & 56.9 & 66.8 \\
3  & 61.9 & 78.8 & 69.4 & 57.3 & 66.9 \\
5  & 62.2 & 78.9 & 69.0 & 57.9 & 67.0 \\
7  & 61.9 & 79.0 & 69.1 & 58.6 & 67.1 \\
9  & 62.2 & 79.0 & 68.7 & 58.8 & 67.2 \\
11 & 60.6 & 78.2 & 69.2 & 57.6 & 66.4 \\\bottomrule
\end{tabular}%
}
\end{table}

\begin{table}[]
\centering
\caption{Results of dropping non-active attention heads as a visual pruning strategy to prune the LLaVA-1.5-7B model. '\# Heads' indicates the number of dropped non-active attention heads per layer used in visual token computation.  }\label{tab:7Bne_head}
\resizebox{0.8\columnwidth}{!}{
\begin{tabular}{cccccc}
\toprule
\# Heads & {GQA} & {VQAv2} & {SQA} & {TextVQA} & {Avg} \\\hline
4 & 62.1 & 78.9 & 68.0 & 58.3 & 66.8 \\
12 & 61.9 & 78.6 & 68.3 & 57.3 & 66.5 \\
20 & 61.6 & 78.1 & 67.7 & 57.2 & 66.1 \\
28 & 60.8 & 77.1 & 67.7 & 56.0 & 65.4 \\\bottomrule
\end{tabular}}
\end{table}

\begin{table}[]
\centering
\caption{Results of applying the layer-dropping strategy to prune the LLaVA-1.5-13B model. \# layers indicates the number of last several layers to drop for visual-related computation.  }\label{tab:7Bne_ld}
\resizebox{0.8\columnwidth}{!}{
\begin{tabular}{cccccc}
\toprule
\# Layers & {GQA} & {VQAv2} & {SQA} & {TextVQA} & Avg \\\hline
4 & 62.4 & 79.1 & 68.7 & 57.8 & 67.0 \\
8 & 62.2 & 79.1 & 68.7 & 58.3 & 67.0 \\
12 & 62.2 & 79.0 & 68.9 & 57.5 & 67.0 \\
16 & 62.1 & 78.7 & 68.4 & 57.0 & 66.6 \\
20 & 61.4 & 78.0 & 68.0 & 55.3 & 65.7 \\
24  & 59.3 & 75.4 & 68.1 & 54.0 & 64.2 \\
28  & 54.5 & 71.1 & 68.4 & 51.6 & 61.4 \\\bottomrule
\end{tabular}}
\end{table}

It can be shown that even without any fine-tuning, our proposed pruning strategies can efficiently reduce the required FLOPs with remaining similar performance compared with the original model. It should be noticed that from the results of pruning InternVL-2.0 models, we can see that \textit{a larger model usually has a larger computation redundancy.} For example, while the use of $44\%$ FLOPs causes a large performance drop for InternVL-2.0-8B model, the InternVL-2.0-26B model still has a comparable performance compared with the original one.  Also, for a given pruning ratio, we should carefully select the pruning strategy. From the results of Qwen2-VL-8B, the use of neighbor-aware attention computation could cause a larger performance drop than the use of non-active attention head dropping. It remains further work to study how to select a better combination of pruning strategies to get a good performance with improved efficiency.

\section{Demo application}
\begin{figure*}
    \includegraphics[width=\linewidth]{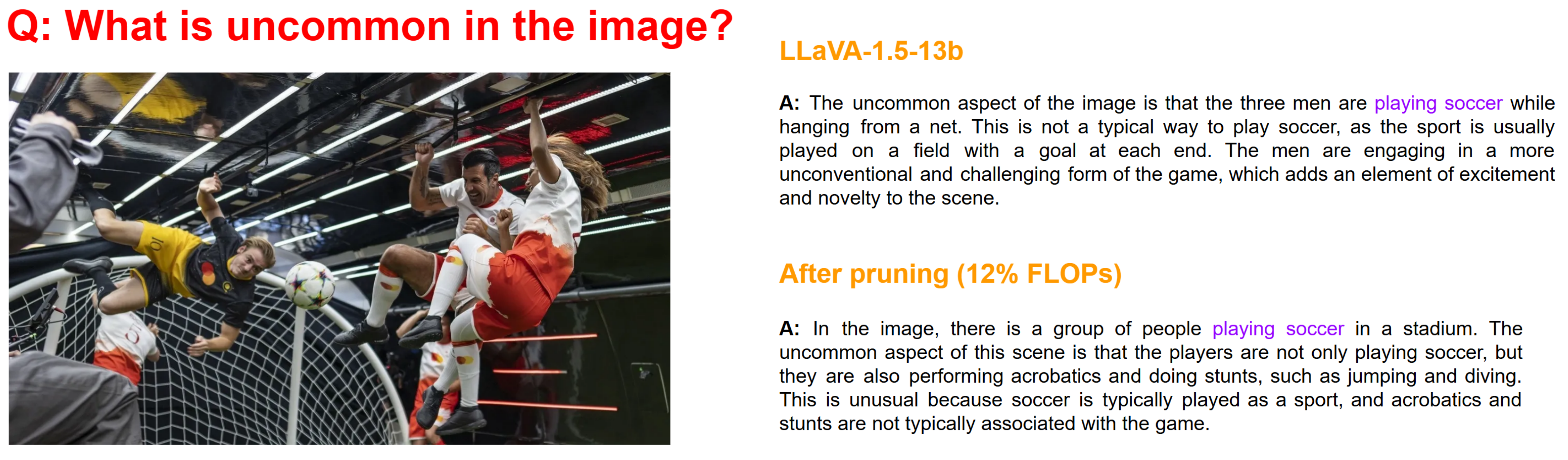}
    \caption{Demo application result \#1.}\label{demo:1}
\end{figure*}

\begin{figure*}
    \includegraphics[width=\linewidth]{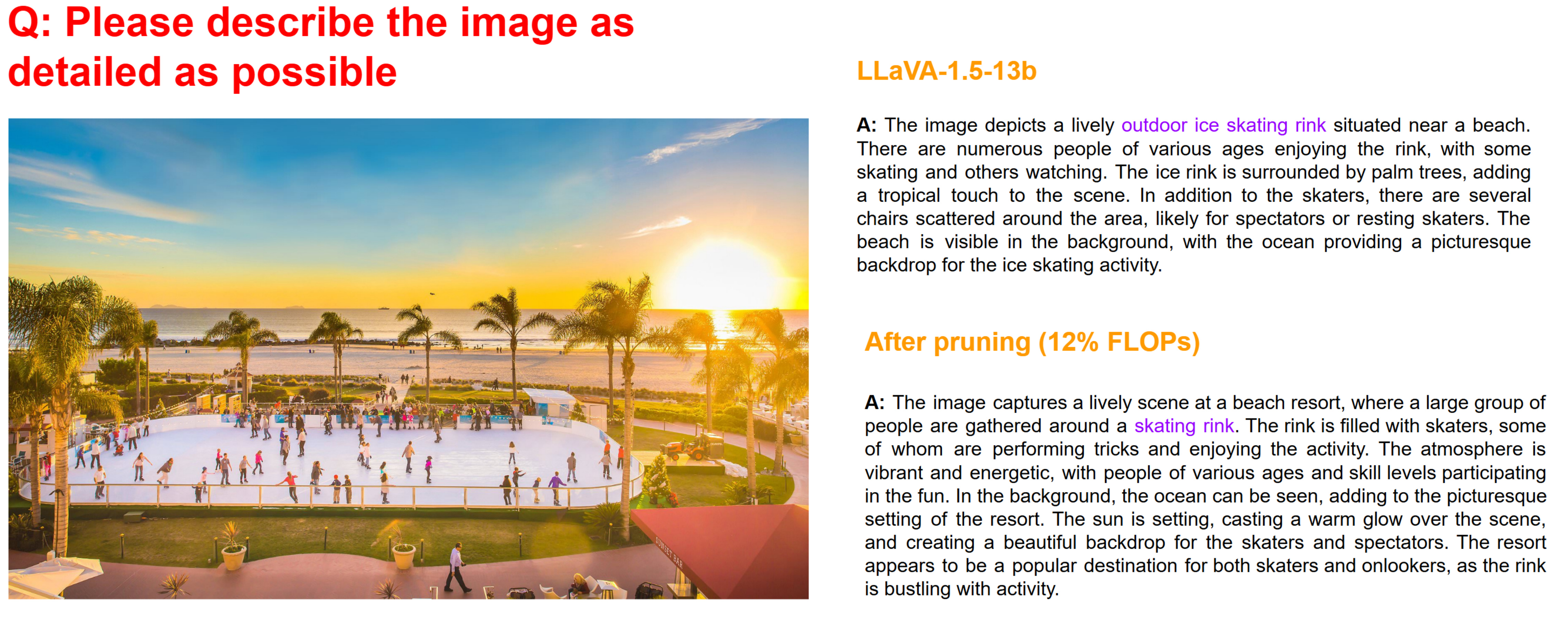}
    \caption{Demo application result \#2.}\label{demo:2}
\end{figure*}

\begin{figure*}
    \includegraphics[width=\linewidth]{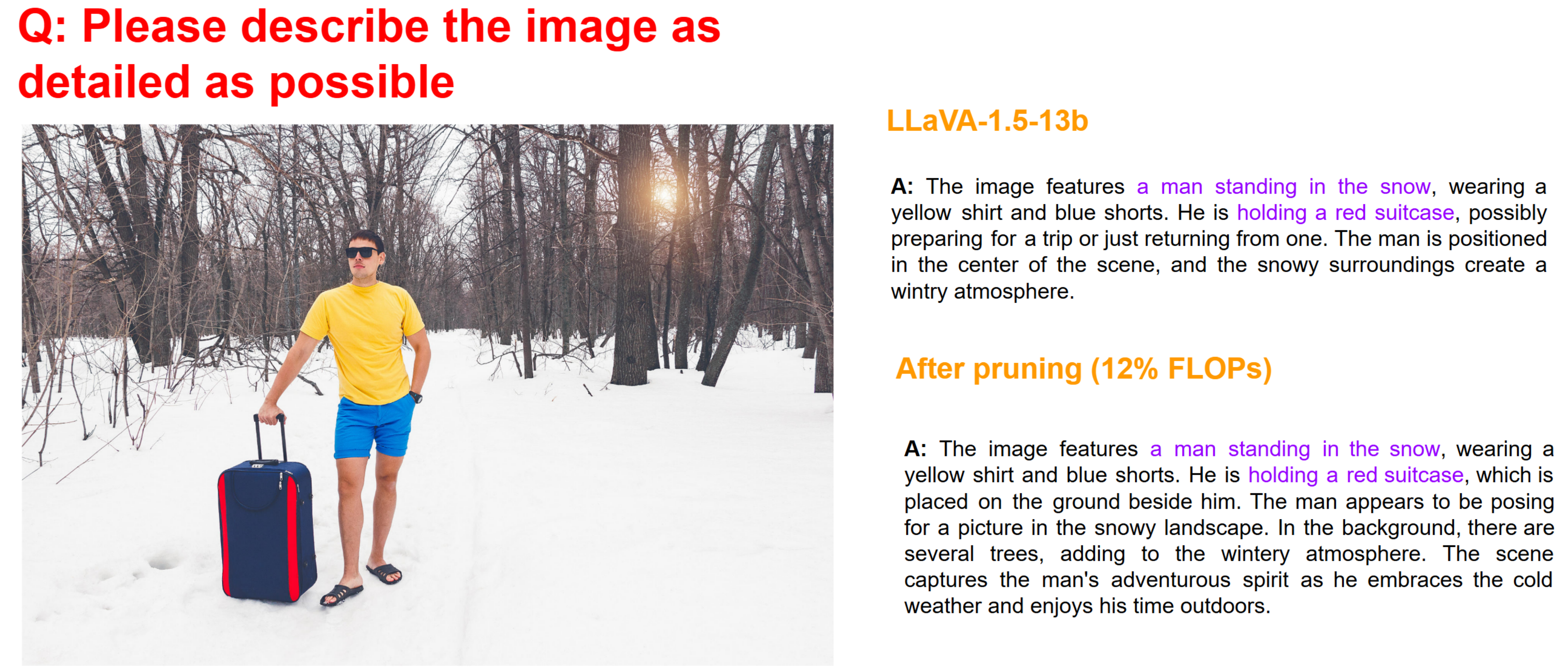}
    \caption{Demo application result \#3.}\label{demo:3}
\end{figure*}

We also present three real-world examples to show the performance of the model (LLaVA-1.5-13B) before and after pruning. The results are depicted in \cref{demo:1}, \cref{demo:2}, and \cref{demo:3}. We highlight the expected keywords in \textcolor{violet}{violet}.  Even with only 12\% of FLOPs allocated to visual-related computation, MLLM demonstrates the ability to accurately capture semantic information. This highlights the significant potential of MLLM visual computation pruning in accelerating real-world applications.

